# DUAL UNET: A NOVEL SIAMESE NETWORK FOR CHANGE DETECTION WITH CASCADE DIFFERENTIAL FUSION

*Kaixuan Jiang, Ja Liu\*, Fang Liu, Wenhua Zhang, Yangguang Liu*

Jiangsu Key Laboratory of Spectral Imaging & Intelligent Sense, Nanjing University of Science and Technology, Nanjing, 210094, China

## ABSTRACT

Change detection (CD) of remote sensing images is to detect the change region by analyzing the difference between two bitemporal images. It is extensively used in land resource planning, natural hazards monitoring and other fields. In our study, we propose a novel Siamese neural network for change detection task, namely Dual-UNet. In contrast to previous individually encoded the bitemporal images, we design an encoder differential-attention module to focus on the spatial difference relationships of pixels. In order to improve the generalization of networks, it computes the attention weights between any pixels between bitemporal images and uses them to engender more discriminating features. In order to improve the feature fusion and avoid gradient vanishing, multi-scale weighted variance map fusion strategy is proposed in the decoding stage. Experiments demonstrate that the proposed approach consistently outperforms the most advanced methods on popular seasonal change detection datasets.

## 1. INTRODUCTION

The object of change detection is designed to identify semantic change information in the same region at different periods. CD technology can be applied in a variety of fields, such as earthquake rescue, urban planning and military detection [1]. Bi-temporal CD is a comprehensive analysis of images, where each point in the image is marked into two groups: changed and unchanged classes. Through analysis, the final difference diagram and CD results can be obtained [2].

The traditional change detection method is performed as follows: (1) image preprocessing, such as geometric alignment and radiation correction; (2) pixel-wise difference generation via comparing the features extracted from the bitemporal images [1]; (3) image segmentation [3] to classify the pixels into changed and unchanged ones. However, such methods highly depend on expert experiences and there are many manually determined parameters which result in difficulty of dealing with large amount data with high accuracy. As a consequence, there are some limitations at the application level [4]. With the rapid growth of deep learning methods, many neural network models and components are used in the field of CD to extract deeper feature representations, which makes it possible to extract feature maps through an end-to-end approach [5]. Although the CD methods have all achieved practical success, there are still some problems. Due to the influence of irrelevant inconsistency such as illumination variations and alignment errors, many existing methods are difficult to accurately capture the spatial local information of images [6]. It leads to uncertainty in pixel edges and misjudgment of the target.

In this paper, an improved U-Net [7] called Dual-UNet, is presented to capture differential information at different dimensions by adding a multi-scale differential attention module in the encoder stage, which improves the ability of the image to extract change information in the shallow feature extraction stage. In addition, a weighted difference fusion mechanism is introduced into the decoding stage for obtaining high-level semantic difference features at different levels of the image and weighting coefficients are added to learn the influence factor of each layer on the difference map. Experiments were performed in the season-varying change detection dataset and the experimental results indicate that the proposed approach model outperforms several other recent methods.

## 2. RELATED WORK

In accordance with the detection strategy, most of the early change detection methods were doing algebraic operations on the image, such as difference operator, log-ratio operator. With the flourishing of deep learning technology, especially deep convolutional neural networks (CNN), learning data representation with multi-level abstraction has been widely used in remote sensing [8]. Long at al. [9] introduced the fully convolutional neural network (FCN), which is an end-to-end

This work was supported in part by the National Nature Science Foundations of China (Grant Nos.61906093 and 61802190), in part by Open Research Fund in 2021 of Jiangsu Key Laboratory of Spectral Imaging & Intelligent Sense (Grant No. JSGP202101), and n part by the Nature Science Foundations of Jiangsu Province, China (Grant No. BK20190451).

\*Corresponding author: Jia Liu. Email: omegaliuj@gmail.com

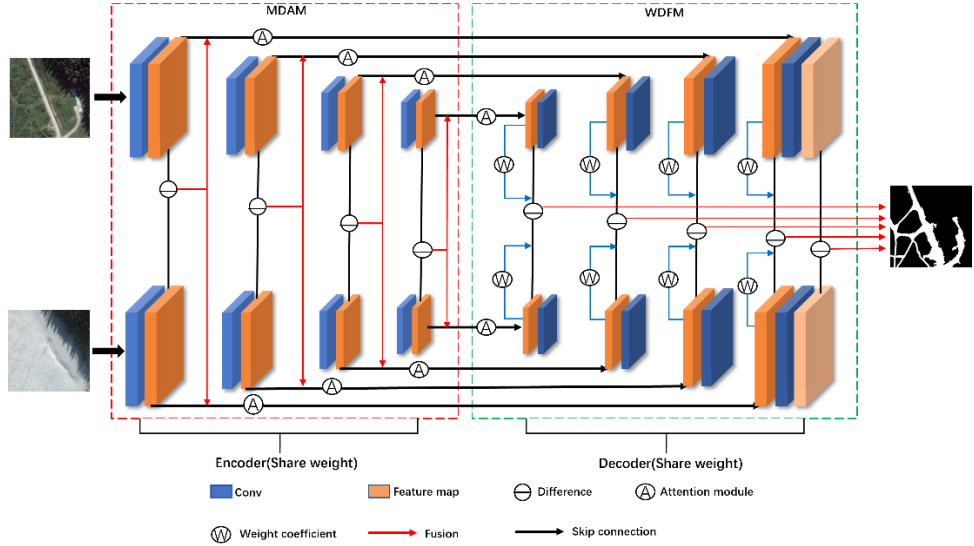

**Fig.1** Flowchart of Dual-UNet architecture

training model. FCN is widely used in change detection tasks, and many variant networks of FCN have been generated afterwards. Especially U-Net, was used to deal the pixel-level CD task. Daudt et al. [10] proposed three U-Net based change detection methodology, including early merge, Siamese concatenation, and Siamese difference. However, the network only uses convolutional layers to obtain change information, which is unable to extract a deep feature map representation. In order to extract deep-level change information, Chen et al. explored STANet, which feeds the feature map into a multi-scale attention module after extracting network features, which can capture spatial relationships at different scales [5][11].

The attentional mechanism is a mimicry of the human visual attention mechanism. It is essentially a resource allocation mechanism that aims to acquire more detailed information and suppress other useless information. Numerous experiments have shown that the attention mechanism has better performance in image denoising, image feature extraction, super-resolution assignments and oppression artifact mitigation.

## 3. PROPOSED METHOD

### 3.1 Dual-Net Structure

We propose a new Siamese U-Net based multiscale differential attention module (MDAM) and weighted difference fusion mechanism module (WDFM). The backbone architecture of this network does not encode dual-temporal images alone, but a bilateral U-Net network with shared weights.

As shown in Fig.1, the network has two modules: MDAM and WDFM. in the MDAM module, the bitemporal images are supplied into the bilateral encoder network separately to catch the multi-scale feature maps. Afterwards, the difference maps are obtained by differencing the feature maps at each scale, and the difference maps are fused with the original feature maps in order to enhance the change regions, and the fused features are processed by the attention mechanism to obtain a more significant change feature map. After that, in the WDFM module, the feature maps generated in the encoder are skip-connected to the feature maps obtained by upsampling in the decoder. The feature maps at each scale are multiplied by a weight matrix $w$, the size of $w$ is the same as the size of the feature maps at each scale, and the channels of the $w$ matrix is 1. The $w$ is used to learn the contribution coefficients of the feature maps at various scales to the final difference map. then, the feature maps at varied scales in the decoder stage are differenced to obtain the multi-scale variance maps. Eventually, the multi-scale difference maps are fused and the final CD results are obtained from the threshold method.

### 3.2 multiscale differential attention module (MDAM)

In the change detection task, since changing objects may have various dimensions, abstracting features from a proper range can efficiently fuse feature maps from different scales, which can better represent the spatial information of the target. Driven by this motivation, we introduce attention mechanisms at different scales in the coding process.

In order to highlight the variation regions, we make differences in the feature maps at each layer of the bilateral encoder to get the difference feature maps at different levels, after which the difference feature maps at each layer are sent to the encoder to fuse with the original feature maps. Finally the fused feature maps are subjected to the self-attention mechanism to focus on the shallow-level difference features, after which they are connected to the decoding stage by skip connections [11].

### 3.3 weighted difference fusion mechanism (WDFM)

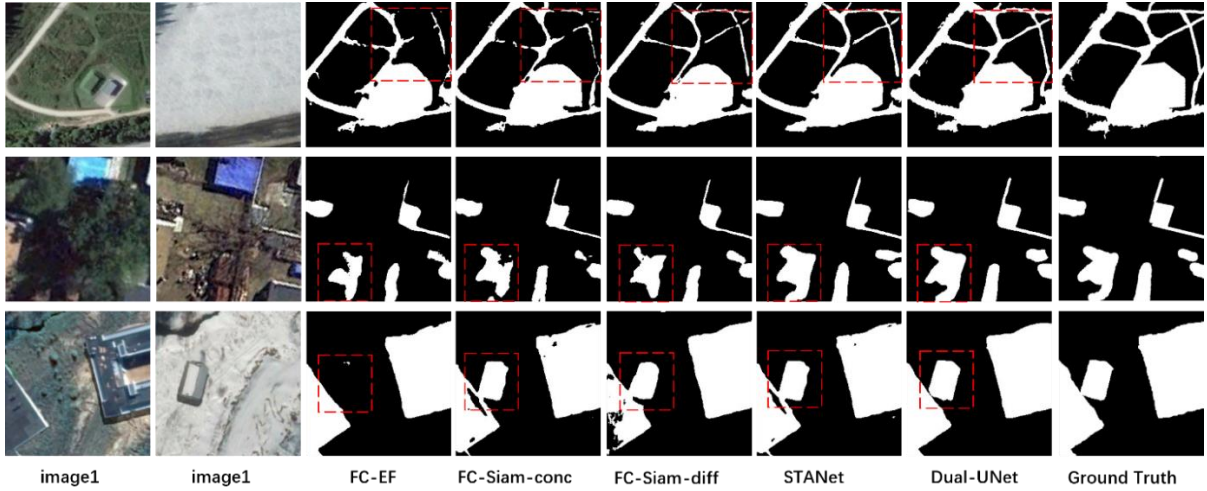

**Fig.2** Change detection examples of different methods on the CDD dataset

In the encoder stage, both Siam-UNet and STANet used to do the difference of the extracted last layer feature maps, and then the final difference maps were obtained by thresholding. This method does not adapt well to the scale variations of different objects. In an attempt to solve the problem, we introduce a weighted difference fusion mechanism, in which we perform difference processing for each layer of the upsampled feature map, and in the meantime, introduce a weight matrix for each layer of the feature map, through which we learn the contribution coefficients of the feature maps at different levels to the final difference map, and the weight coefficients can be used to focus on the distinguishing features of different objects at different levels. Finally, the level-by-level difference maps obtained from the decoder stage are fused to obtain the final difference maps.

### 3.4 Loss Function

For CD, in general the changing samples are much smaller than the unchanging samples, which makes the network prone to bias during training. To decrease the influence of loss imbalance, we use batch balancing contrast loss, which aims to modify the weights of different loss terms of the loss function to reach a balancing effect. Given a batch of samples $(X^{*(1)}, X^{*(2)}, M^*)$, $X^{*(1)}$, $X^{*(2)} \in \mathbb{R}^{B \times 3 \times H_0 \times W_0}$, $M^* \in \mathbb{R}^{B \times H_0 \times W_0}$, then we can get the distance map $D^* \in \mathbb{R}^{B \times H_0 \times W_0}$, where $B$ is the batch size of samples, $H_0$ and $W_0$ denote height and width of feature map. $M^*$ is a group of binary label, where 1 represents change value and 0 denotes no change value. Batch-balanced contrastive loss (BCL) is described as follows:

$$L(D^*, M^*) = \frac{1}{2}\frac{1}{p_u}\sum_{b,i,j}(1-M^*_{b,i,j})D^*_{b,i,j} \\ + \frac{1}{2}\frac{1}{p_c}\sum_{b,i,j}M^*_{b,i,j}Max(0, m-D^*_{b,i,j}) \quad (1)$$

where $b$, $i$ and $j$ are factors that batch, height and width respectively. $m$ denotes the boundary value to limit the size of the changed pixel pairs, which is set to 2 in the experiment, where $p_u$, $p_c$ are the number of unchanged pixel pairs and changed pixel pairs, respectively, they are calculated from the label values of the corresponding categories:

$$p_u = \sum_{b,i,j} 1 - M^*_{b,i,j}, \quad p_c = \sum_{b,i,j} M^*_{b,i,j} \quad (2)$$

## 4. EXPERIMENTAL RESULTS

### 4.1 Datasets

The season-varying change detection dataset (CDD) includes 16000 realistic pairs of the Google Earth image of the changing seasons with pixel-by-pixel change detection labels, containing 3000 testing samples, 3000 validation samples and 10000 training samples. Each image is approximately 256 × 256 pixels in size. CDD can provide information not only on common object changes such as buildings and land, but also on changes in many detailed objects such as cars and roads. Based on this data division we designed two sets of experiments to completely demonstrate the effectiveness of Dual-UNet.

### 4.2 Experimental Design

To fully validate the efficiency of Dual-UNet, we have designed two types of experiments. The same hyperparameter settings are used for all comparisons.

*4.2.1. Ablation Investigation*

Ablation research was initially carried on the CDD dataset. To verify the effectiveness of the multiscale differential attention module and weighted difference fusion mechanism, we designed one baseline method.

(1) Base: Siamese Unet architecture with shared weights
(2) Proposed1: Base + MDAM

(3) Proposed2: Base + MDAM+WDFM

We verified the validity of the spatiotemporal module by we verified the validity of the spatiotemporal module by comparing BASE, MDAM and WDFM. Table 1 shows the baseline ablation study and its variation on the CDD test dataset. The Precision, Recall and IoU associated with the category of variation were calculated to evaluate the capabilities of our method. We can see that MDAM and WDFM show a significant improvement over baseline. The IoU of MDAM is improved by 1.85 points compared to the baseline. In addition, our multi-scale attentional design (WDFM) has significantly improved performance with a 4.62 point improvement in IoU compared to MDAM.

Table 1. Ablation study of MDAM and WDFM on CDD

| Method | Precision(%) | Recall(%) | IoU(%) |
|---|---|---|---|
| Base | 91.99 | 92.48 | 85.59 |
| Proposed1 | 94.29 | 92.32 | 87.44 |
| Proposed2 | **97.46** | **94.31** | **92.06** |

*4.2.2. Comparative Experiments*

We also evaluated our presented method on the CDD dataset, and compared with several advanced bitemporal image CD methods: FC-EF, FC-Siam-diff, FC-Siam-conc [10] and STANet [6]. In FC-EF [10], the EF architecture connects the two patches together, and then through the UNET network structure. In FC-Siam-diff, to detect the change area of image. FC-Siam-diff concatenates the difference in the feature map extracted in the encoding stage to obtain more detailed change information. In FC-Siam-conc, it is a concatenated extension of FC-EF model. The encoder has been divided into two streams with the same structure with shared weights. Each picture is fed into the same stream, and two skip connections are connected in the decoding step, each from an encoded stream. In STANet [6], the network uses Resnet-18 for its feature extractor, and introduces the convolution self-attention module into the feature extraction stage. Finally, the CD results is obtained by counting the distance mapping between features maps. Table2 indicates the comparison of different methods on CDD datasets. We have observed that our proposed method was consistently superior to other advance CD methods in F1-score. Figure 2 shows examples of different CD methods on the CDD dataset. As can be seen from the diagram, The red dashed box in the figure indicates the results of differences captured by different methods. We can see that our proposed approach can more effectively capture the differences of buildings and get smoother results.

Table 2. COMPARATIVE EXPERIMENTS ON CDD

| Methods | Precision(%) | Recall(%) | F1(%) | IoU(%) |
|---|---|---|---|---|
| FC-EF | 86.15 | 54.51 | 66.77 | 50.12 |
| FC-Siam-diff | 82.67 | 71.47 | 76.65 | 62.16 |
| FC-Siam-conc | 87.47 | 71.58 | 78.73 | 64.93 |
| STANet | 94.88 | 89.72 | 92.23 | 85.58 |
| Dual-UNet | **97.46** | **94.31** | **95.86** | **92.06** |

## 5. CONCLUSION

In this paper, we have proposed a Dual-UNet for remote sensing binary CD. Dual-UNet consists of two concurrent encoders and decoders with shared weights, the former for extracting multi-scale features and the latter for decoding change information by incorporating different differential feature maps. The ablation experiments proved the effectiveness of our proposed MDAM and WDFM module, which can obtain the correspondence between different features and intermediate semantic change maps at different levels. Both of them have good results in the tasks of image denoising, compression artifact reduction. The experimental outcomes reveal that our proposed network can well mitigate the false detection caused by bitemporal image alignment errors. On the CDD dataset, our proposed method outperforms several other advanced remote sensing CD methods. In addition, by extracting multi-scale attentional features and fusing the weighted differential feature maps, our method is more robust to scale and color variations in bitemporal images.

## 6. REFERENCES

[1] M. Gong, L. Su, H. Li, and J. Liu, "A Survey on Change Detection in Synthetic Aperture Radar Imagery," J. Comput. Res. Dev., 2016.

[2] H. M. Keshk and X.-C. Yin, "Change Detection in SAR Images Based on Deep Learning," Int. J. Aeronaut. Space Sci., vol. 21, no. 2, Art. no. 2, 2020.

[3] H. Sui, W. Feng, L. I. Wenzhuo, K. Sun, and X. U. Chuan, "Review of Change Detection Methods for Multi-temporal Remote Sensing Imagery," vol. 43, no. 12, Art. no. 12, 2018.

[4] J. Liu, M. Gong, A. K. Qin, and K. C. Tan, "Bipartite Differential Neural Network for Unsupervised Image Change Detection," IEEE Trans. Neural Netw. Learn. Syst., vol. PP, no. 99, Art. no. 99, 2019.

[5] J. Zhao, M. Gong, J. Liu, and L. Jiao, "Deep learning to classify difference image for image change detection," in 2014 International Joint Conference on Neural Networks (IJCNN), Beijing, China, Jul. 2014.

[6] H. Chen and Z. Shi, "A Spatial-Temporal Attention-Based Method and a New Dataset for Remote Sensing Image Change Detection," Remote Sens., vol. 12, no. 10, Art. no. 10, 2020.

[7] O. Ronneberger, P. Fischer, and T. Brox, "U-Net: Convolutional Networks for Biomedical Image Segmentation," Springer Int. Publ., 2015.

[8] Y. T. ZHANG Xinlong CHEN Xiuwan, LI Fei, "Change Detection Method for High Resolution Remote Sensing Images Using Deep Learning," Acta Geod. Cartogr. Sin., vol. 46, no. 8, Art. no. 8, 2017.

[9] E. Shelhamer, J. Long, and T. Darrell, "Fully Convolutional Networks for Semantic Segmentation," IEEE Trans. Pattern Anal. Mach. Intell., vol. 39, no. 4, Art. no. 4, Apr. 2017.

[10] R. Caye Daudt, B. Le Saux, and A. Boulch, "Fully Convolutional Siamese Networks for Change Detection," in 2018 25th IEEE International Conference on Image Processing (ICIP), Athens, Oct. 2018.

[11] Vaswani, Ashish, et al. "Attention is all you need." Advances in neural information processing systems. 2017.